\documentclass[review]{elsarticle}

\usepackage{lineno,hyperref}
\modulolinenumbers[5]

\journal{Annals of Physics}

\begin{document}

\begin{frontmatter}

\title{Order-Fractal transition in abstract paintings}

\author{E.M. de la Calleja*}
\address{Instituto de F\'{i}sica, Universidade Federal do Rio Grande do Sul, Caixa Postal 15051, 91501-970, Porto Alegre, RS, Brazil.}
\cortext[]{elsama79@gmail.com. E.M.D.C. Bolsista do CNPq - Brazil.}

\author{F. Cervantes}
\address{Department of Applied Physics, CINVESTAV-IPN, Carr. Antigua a Progreso km.6, Cordemex, C.P.97310, M\'erida, Yucat\'an, M\'exico.}

\author{J. de la Calleja}
\address{Department of Informatics, Universidad Polit\'ecnica de Puebla, $72640$, M\'exico.}




\begin{abstract}
We report the degree of order of twenty-two Jackson Pollock's paintings using \emph{Hausdorff-Besicovitch fractal dimension}.
Through the maximum value of each multi-fractal spectrum, the artworks are classify by the year in which they were painted.
It has been reported that Pollock's paintings are fractal and it increased on his latest works.
However our results show that fractal dimension of the paintings are on a range of fractal dimension with values close to two.
We identify this behavior as a fractal-order transition.
Based on the study of disorder-order transition in physical systems, we interpreted the fractal-order transition through its dark paint strokes in Pollocks' paintings, as structured lines following a power law measured by fractal dimension.
We obtain self-similarity in some specific Pollock's paintings, that reveal an important dependence on the scale of observation.
We also characterize by its fractal spectrum, the called \emph{Teri's Find}.
We obtained similar spectrums between \emph{Teri's Find} and \emph{Number 5} from Pollock, suggesting that fractal dimension cannot be completely rejected as a quantitative parameter to authenticate this kind of artworks.
\end{abstract}


\end{frontmatter}

\section{Introduction}
Fractality is present in many objects in nature, in structures generated by mathematical algorithms, in spacial interaction among populations, on distributions of particles in amorphous solids or in particles configurations created by computer simulations~\cite{Barnsley,Benoit,Vicsek,Meakin,Spehar1}.
It means that, we can measure fractal characteristics in a wide variety of two-dimension digital images.
From the point of view of the analysis of physical systems, it is possible to identify fractal characteristics in Pollock's paintings~\cite{TayNat99,Shamir}

Since R. P. Taylor \emph{et al.} presents its famous result, indicating that all the Jackson Pollock's paintings are fractals~\cite{TayNat99},
many reports have been published confirming or questioning the fractal characteristic of abstract artworks~\cite{Spehar,TayMic03,Coddington,MureikaChaos,Jones,Jones1}.

Based on fractal dimension, Taylor presented five criteria to describe the construction processes of the colored layers on the Pollock's canvases~\cite{TayGuz07}.
He additionally concluded that it is possible to authenticate the paintings using fractal dimension~\cite{TayNat99,TayGuz07,Tay10}.
Taylor's criteria have been tested by different methods, and there are many reports where conclude that it is not enough use fractal dimension to authenticate any drip painting~\cite{Coddington,MureikaChaos,Jones,Jones1,Rockmoreauthen,Mureikamulti}.

However, the objections on these works have been addressed by others~\cite{Irfan,Hany}, leaving the fractal dimension as a rigorous measure to characterize in a good way the Jackson Pollock's paintings; used as an important parameter on the authentication process, demonstrating its useful to improve the analysis of complex abstract art~\cite{Coddington,Rockmoreauthen}.

Pollock's paintings were created by dripping, pouring, splashing or peeling layers of paint of different colors on canvas placed on the floor.
Paint strokes above each others are painted, until the canvas is totally covered, or at least in large part of it.
Pollock argued that he had control of the splattering or dripping over the paints, denying the accidental paintings~\cite{Book-Pollock,film}.
This argument suggest that he was aware about the action in the tact, the movements to control the paint strokes, the flow of each paint,
the velocity and the rhythm in his creative processes.
For many researches it is surprising considering the apparent complexity of the strokes.

The visual complexity of the linked paint strokes results by the method described above apparently poses a high degree of complexity, taking a definition that it is directly proportional to a high number of linked lines superimposed.
For the case of Pollock's paintings, we found that it is not completely true.
If Pollock spills paint lines, adding one by one, in a particular way which only the artist knows, we can identify that the process perhaps is not completely in a random way.
Certainly the creative process is unique, complex, and unrepeatable, specially in the case of Pollock.

In this report we compare the physical complexity found in two-dimensional images from experimental phenomena and we used the same methodology to quantify the degree of order in physical structures.
The paint strokes distribution on Pollock's paintings has been compared to natural objects, however we related them to some fibrilar aggregation~\cite{Jordens} and particle aggregation processes~\cite{Calleja1,Calleja3,Cervantes,Gonzales}.
The distribution of paint strokes reflects some degree of disorder, but we assumed a high degree of order if the fractal dimension is close to two,
According to previous results of treatment of physical images that describe order evolution in the framework of supercooled liquids~\cite{Calleja1}, granular materials~\cite{Gonzales} and magnetic properties~\cite{Carrillo2}, we believe that the structure formed by paint strokes posses a high degree or order.

Our results suggest an increasing order of the paint strokes distribution from a particular array of his paintings (See Table~\ref{table-1}), also our results indicate that the fractal property of Pollock's paintings is in a specific range of values.
It has been discussed that fractal dimension can not be used as a quantitative parameter to authenticate abstract art.
However, we point out the case of \emph{Teri's Find}, which has been the famous case of discussion about the authentication process and has been evidenced the methodology by art historians~\cite{Biro} to carry out the complex process of authenticating artworks. We tested the paint and interesting results between this painting and \emph{Number 5} from Pollock, are discussed here.

\section{Multi-fractal spectrum}
Fractality is a geometrical, topological, structural and beauty property present in many natural, physics or simulated systems.
This property can be recognized and calculated in many two-dimensional structures through digital images~\cite{Barnsley,Benoit,Vicsek,Gouyet}.
In addition, there has been reported that fractal structures are result of kinetic aggregation or/and reaction processes~\cite{Meakin,Schmitt,Ben-Jacob,Naito}.

The multi-fractal spectrum has been used as a measure of all the local fractal dimensions coexisting in spatial structures of a wide variety of physical, chemical and biological systems.
This measure of complexity has been employed to characterize the structural transition of rheological fluids~\cite{Calleja1,Cervantes,Calleja2}, granular materials~\cite{Gonzales,Pusey}, magnetic wall domains in boracite~\cite{Carrillo2} and other complex systems~\cite{Suzuki,Gonzalez2}.

The width of multi-fractal spectrum, calculated by the well-known box-counting method, is affected by the treatment applied on digital images~\cite{Abry}.
However there has been proposed different methods to analyze and treat complex digital images~\cite{Amirshahi,Kovalevsky,Stork} to obtain a dependable measure.
All these methods confirm the validity of many free computer programs to analyze, by fractal dimension, a wide variety of digital images from a extensive variety of systems~\cite{Gouyet,Redies,Falconer,Lopez}, including the abstract artworks~\cite{MureikaChaos,Tay10,Mureikamulti,Abry,Sarkar}.

The multi-fractal spectrum generated by an infinite set of dimension measures the scaling structure as a function of the local pattern density.
This give us information about the structural properties at different scales~\cite{Halsey} and also describes the generalized dimensions~\cite{Hentschel}.

The standard scaling relation to relate the number of boxes to cover the set $N(\varepsilon)$ of size $\varepsilon$ is
\begin{equation}
N(\varepsilon)\sim\varepsilon^{-D_{Q}}
\end{equation}
where $\varepsilon$ acquired successively smaller values of length until the minimum value of $\varepsilon_{0}$.
This defines the fractal dimension as
\begin{equation}
D_{Q}=\lim_{\varepsilon \rightarrow 0}\frac{\ln N(\varepsilon)}{\ln (\varepsilon_{0}/\varepsilon)}
\end{equation}

To calculate all the local fractal dimension~\cite{Halsey,Ott}, we used the generalized box counting dimension~\cite{Hentschel,Feigenbaum,Procaccia} defined as
\begin{equation}
D_{Q}=\frac{1}{1-Q}\lim_{\varepsilon\rightarrow 0}\frac{ln I(Q,\varepsilon)}{ln (\varepsilon_{0}/\varepsilon)}
\end{equation}
where
\begin{equation}
I(Q,\varepsilon)=\sum_{i=1}^{N(\varepsilon)}[P_{(i,Q)}]^{Q}
\end{equation}

We are taking into account the scaling exponent defined by Halsey et al.~\cite{Halsey} as $P_{i,Q}^{Q}\sim\varepsilon_{i}^{\alpha Q}$ where $\alpha$ can take a width range of values measuring different regions of the set.
When \emph{Q=0} the generalized fractal dimension represents the classic fractal dimension~\cite{TayMic03}.

As the digital image in a gray scale is divided into pieces of size $\varepsilon$, it suggested that the number of times that $\alpha$ in $P_{i,q}$ takes a value between $\alpha'$ and $d\alpha'$ defined as $d\alpha'\rho(\alpha')\varepsilon^{-f(\alpha')}$ where $f(\alpha')$ is a continuous function.

As \emph{Q} represents different scaling indices, we can define
\begin{equation}
I(Q,\varepsilon)=\sum_{i=1}^{N(\varepsilon)}[P_{(i,Q)}]^{Q}=\int d\alpha'\rho(\alpha')\varepsilon^{-f(\alpha')+Q\alpha'}
\end{equation}
$\alpha_{i}$ is the  Lipschitz-H\"{o}lder exponent, which characterizes the singularity strength in the \emph{ith} box.
The factor $\alpha_{i}$ quantifies the distribution of complexity in an spatial location.

The multi-fractal spectrum is a set of overlapping self-similar configurations.
In that way, we used the scaling relationship taking into account $f(\alpha)$ as a function to cover a length scales of observations.
Defining the number of boxes as a function of the Lipschitz-H\"{o}lder exponent $N(\alpha)$, can be related to the box size $\varepsilon$ as
\begin{equation}
N(\alpha)\sim\varepsilon^{-f(\alpha)}\label{eq1}
\end{equation}

The multi-fractal spectrum shows a line of consecutive points for $Q\geq0$ that starts on the left side of the spectrum climbing up to the maximum value.
The values for $Q\leq0$, are dotted on the right side of the spectrum descending until $Q=-10$, which is the minimum value for \emph{Q}.
The maximum value for the generalized dimension corresponds to $Q=0$, which correspond to the box counting dimension.
To obtain the multi-fractal spectrum we use the plugin \emph{FracLac} for ImageJ~\cite{FracLac}.
We select the case of $D_{f}=D_{Q=0}$ as the parameter to quantify the order in the digital images of Pollock's paintings.
In the plugin we select four grid positions that cover the total image.
We apply a gray scale differential option to measure de fractal dimension, and the mode default sampling sizes was selected to plot the values of the spectrum.
We define the smallest sampling element on $30$ px and $100\%$ as the maximum area on analysis of each image.
In this report we present the left side of the spectrum, taking into account the correspondence with thermodynamic formalism~\cite{Chhabra,Chhabra-1}

\section{Pollock's Multi-fractal Spectrum}
We selected twenty-two amazing Pollock's paintings, painted during the called "dripping period".
This selection was done according to the distribution of darkness paint strokes on front, a reduce number of colored layers and its light background.
All the paintings are classified into the movement called {\emph abstract expressionism}.
In table (\ref{table-1}) is presented the list of the selected Pollock's paintings.

\begin{table}
 \caption{Pollock's paintings\label{table-1}}
 \begin{tabular}{llll}
Number & Year & Painting & Size (cm) \\
 1 & 1946 & Free Form~\cite{Pollock1} & 49x36 \\
 2 & 1947 & Lucifer~\cite{Pollock} & 267.9x104.1  \\
 3 & 1947 & Cathedral~\cite{Pollock3} & 89x181.6  \\
 4 & 1947 & Enchanted Forest~\cite{Pollock7} & 221.3x114.6  \\
 5 & 1947 & Reflection of the Big dipper~\cite{Pollock} & 111x91.5  \\
 6 & 1947 & Undulating Paths~\cite{Pollock} & 114x86  \\
 7 & 1947-1950 & Number 19~\cite{Pollock5} & 78.4x57.4  \\
 8 & 1948 & Summertime: Number 9A~\cite{Pollocka} & 84.8x555  \\
 9 & 1948 & Number 26A; Black and White 1948~\cite{Pollock4} & 208x121.7  \\
 10 & 1948 & Number 23~\cite{Pollock2} & 575x784  \\
 11 & 1948 & Number 4(Gray and Red)~\cite{Pollock3} & 58x79 \\
 12 & 1948 & Silver over Black,White,Yellow and Red~\cite{Pollock} & 61x80 \\
 13 & 1948 & Number 14 Gray~\cite{Pollock} & 57x78.5 \\
 14 & 1949 & Number 10~\cite{Pollock3} & 46.04x272.41 \\
 15 & 1950 & Autumn Rhythm; Number 30~\cite{Pollock} & 525.8x266.7 \\
 16 & 1950 & Number 32~\cite{Pollock} & 457.5x269 \\
 17 & 1950 & Number 29~\cite{Pollock3} & 182.9x121.9 \\
 18 & 1950 & Number 18~\cite{Pollock7} & 56x56.7 \\
 19 & 1951 & In Echo: Number 25~\cite{Undulating} & 233.4x218.4 \\
 20 & 1912-1956 & No. 15, 1950~\cite{Pollock6} & 55.88x55.88  \\
 21 & 1951 & Untitled,ink on Japanese paper~\cite{Pollock9} & 62.9x100.3  \\
 22 & 1951 & Untitled~\cite{Pollock8} & 63.5x98.4 \\
\end{tabular}
\end{table}

The multi-fractal spectrum is formed by consecutive points on the left side that grows up until the maximum value, represent all the localized fractal dimensions by different box sizes that means different scales of measure ($\varepsilon$).
All the curves of the multi-fractal spectrum present similar width and length that grows from $\alpha=1.6$.

\begin{figure}
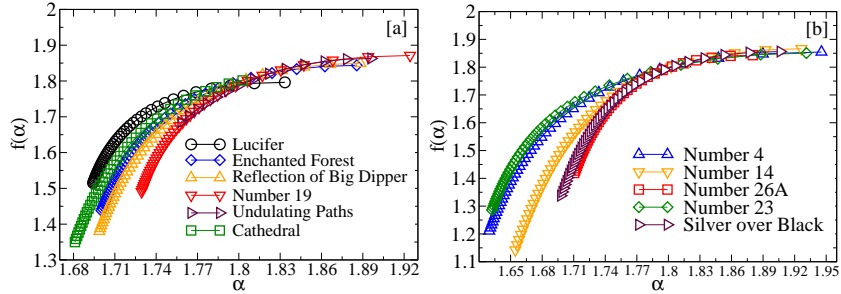

\begin{center}
 \includegraphics[width=0.45\textwidth]{F-1a.eps}
   \includegraphics[width=0.45\textwidth]{F-1b.eps}
\caption{In (a) is shown the left side of the multi-fractal spectrum of Pollock's paintings selected from $1947$.
In (b) are show the corresponding multi-fractal spectrum of Pollock's canvas painted in $1948$.\label{47-48}}
\end{center}
\end{figure}

Fig.(\ref{47-48})\emph{(a)} shows the left side of the multi-fractal spectrum for the selected Pollock's artworks painted in $1947$.
\emph{(7)Number 19} reached the maximum value on the generalized fractal dimension, while \emph{(2)Lucifer} reached the lowest one of this group.
In Fig.(\ref{47-48})\emph{(b)} is presented the spectrums for the corresponding selected paintings from $1948$.

In the group of paintings from $1948$, \emph{(13)Number 14} obtains the highest value of $f(\alpha)$ among them, while \emph{(9)Number 26A} gets the lowest.
It is outstanding the case of \emph{(6)Undulating Paths} where the number of local fractal dimensions is less than all others. Evidently the fractality of Pollock's paintings is unquestionably, however, it is interesting the similarity between all the spectrums, taking into consideration that all the paintings are different, created in different moments and circumstances.

\begin{figure}
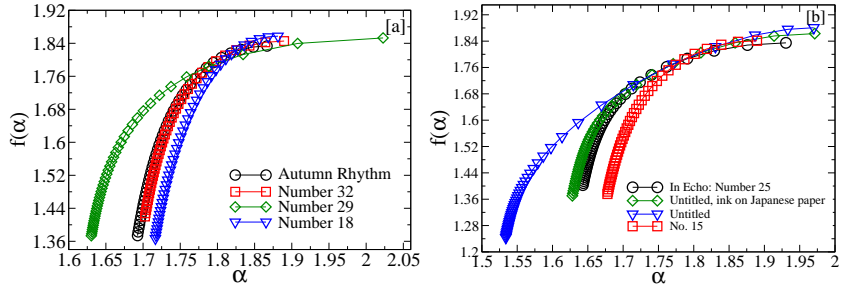

\begin{center}
 \includegraphics[width=0.45\textwidth]{F-2a.eps}
   \includegraphics[width=0.45\textwidth]{F-2b.eps}
\caption{The figure show the left side of the multi-fractal spectrum of Pollock's paintings from $1950$ in (a), and from $1951$ in (b).\label{50-51}}
\end{center}
\end{figure}

Fig.(\ref{50-51}) is shown the left side of the multi-fractal spectrum of the Pollock's artworks painted in $1950$ and $1951$ respectively.
In Fig.(\ref{50-51})\emph{(a)} a similar behavior of the spectrums between the paintings \emph{(15)Autumn Rhythm}, \emph{(16)Number 32} and \emph{(18)Number 18} is observed.
It is possible to recognize similar paint strokes distributed on these canvases, and clear differences with the painting \emph{(17)Number 29}.
The Pollock's paintings created in these years correspond to a period where the technique developed by the artist, was completely dominated.
However they exhibit a wide variety of type paint strokes, very different movements of the paint are visually recognizable and the strokes look so simple on some of them, for example in the cases of \emph{(17)Number 29}, or very complex in other as \emph{(15)Autumn Rhythm}.

In Fig.(\ref{50-51})\emph{(b)} is presented the behavior of the left side of the spectrums of the selected Pollock's artworks painted during $1951$.
We can distinguish the maximum values of fractal dimension for each one.
The painting called \emph{(22)Untitled} reaches the higher value of $f(\alpha)$, while \emph{(19)In Echho: number 25} obtains the minimum one.
The shape of the spectrums with exception of \emph{(22)Untitled}, are similar, obtaining approximately the same number of dimensions.
It notice that the scale for the paintings from $1950$ begins in $\alpha>1.62$ and the spectrum from the paintings from $1951$ arise from $\alpha>1.5$.
It represent differences on scale of observation between the groups of paintings where the complex behavior it shows on the spectrum.

\begin{figure}
\begin{center}
 \includegraphics[width=0.98\textwidth]{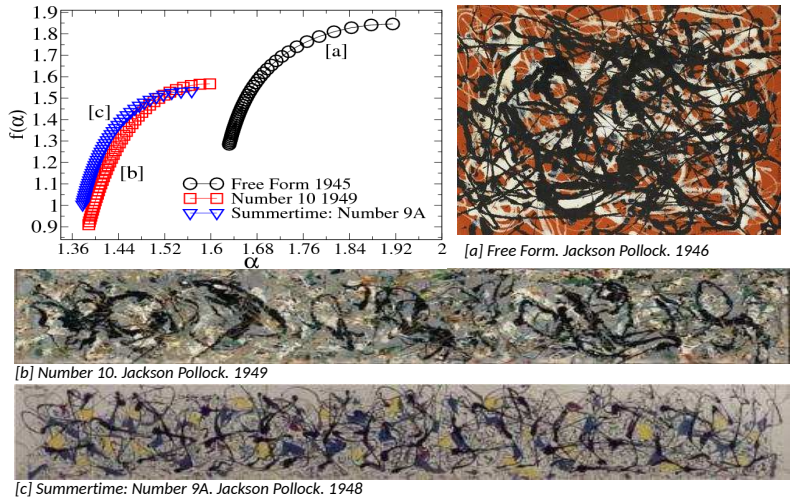}
\caption{The figure shows the left side of the multi-fractal spectrum and a photograph of the Pollock's artworks: \emph{Free Form} painted in $1946$, \emph{Summertime: Number 9A} painted in $1948$ and \emph{Number 10} painted in $1949$.\label{46-49}}
\end{center}
\end{figure}

In Fig.(\ref{46-49}) we show the left side of the multi-fractal spectrum of the paintings \emph{(1)Free Form} painted in $1946$, \emph{(8)Summertime: Number 9A} painted in $1948$ and \emph{(14)Number 10} painted in $1949$.
We can notice a notable difference between the spectrums obtained for these three paintings.
The \emph{(14)Number 10} spectrum grows for $\alpha>1.6$ while in the case of \emph{Free Form} and \emph{(8)Summertime: Number 9A} grows arise from $\alpha>1.5$.

According to equation (\ref{eq1}) the local fractal dimension is manifested for small values of $\alpha$ in comparison with the values for the other paintings presented in Fig.(\ref{47-48}) and Fig.(\ref{50-51}).

The two large paintings: \emph{(14)Number 10} and \emph{(8)Summertime: Number 9A}, present similar number of local fractal dimensions and its spectrums grow from very close $\alpha$ exponent, reach very close maximum values.
There is a notable difference of paint strokes between these paintings and \emph{(1)Free Form} which has major density of paint strokes reflected in the magnitude of scale of $\alpha$.

\section{Order in Pollock's paintings}
The paint strokes on Pollock's canvases are apparently distributed randomly.
However the maximum value of the multi-fractal spectrums presented above, can be associated with different degree of order~\cite{Calleja1,Cervantes,Gonzales,Carrillo2}.
We obtained for an specific array of paintings listed in table (\ref{table-1}), the evolution of the maximum value on the multi-fractal spectrum for each paint, corresponding to the Hausdorf-Besicovich dimension when $Q=0$.

Based on the characterization by fractal dimension of liquid-solid transitions~\cite{Calleja1,Calleja3} we identify that a value of fractal dimension close to two, represents an ordered structure, on the contrary represent a disordered one.
On the case of Pollock's paintings, we measure the linked paint strokes and we obtain, according to the sequence of paintings listed on table (\ref{table-1}) a fractal-order transition.

\begin{figure}
\begin{center}
 \includegraphics[width=0.95\textwidth]{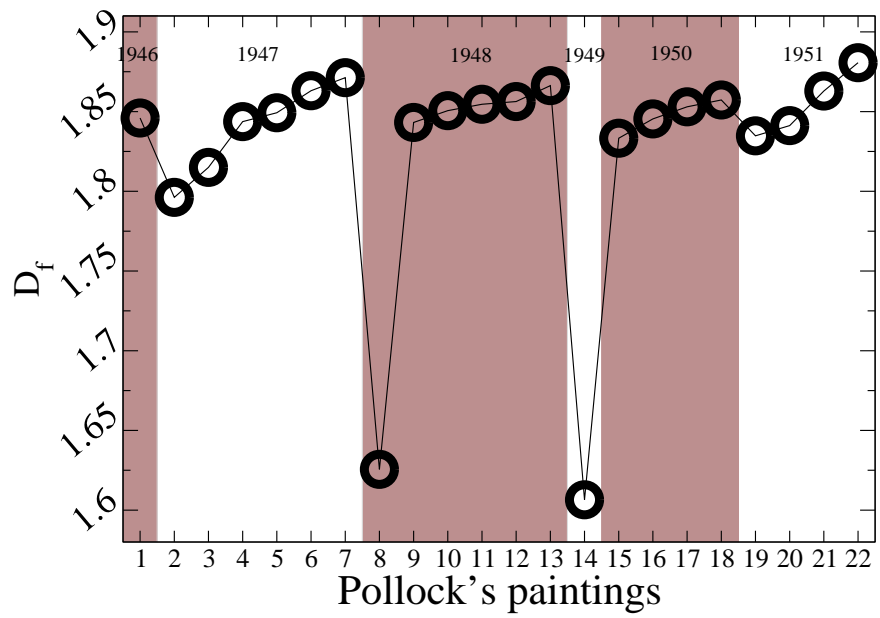}
\caption{The figure shows the evolution over the time of the fractal dimension of Pollock's paintings (following the list on table (\ref{table-1})). The paintings have different complexity within a range of values. The figure show the fractal-order transition for Pollock's paintings grouped by year in which they were painted.\label{FPA}}
\end{center}
\end{figure}

Fig.(\ref{FPA}) shows the evolution of the fractal dimension as a function of the selected Pollock's paintings grouped by year.
We can observe that fractal dimension increases, approaching to a dimension close to two, and it happens in all the groups of paintings.
We can speculate that, if Jackson Pollock paints his artwork in the sequence on table (\ref{table-1}), he added more drip lines that improve the homogenous distribution of them and it is reflecting on the value of the fractal dimension.

For example, on paintings of $1947$, we observe that \emph{(2)Lucifer} obtained the smallest fractal dimension than all and \emph{(7)Number 19} obtain the highest.
This is understood as a result of the distribution of darkness paint strokes that construct a disorder structure on the canvas and also a difference on the density of paint strokes on each painting.
Visually can be see that \emph{(2)Lucifer} look very similar to \emph{(7)Number 19}, however there is high difference on $D_{f}$ of both paintings.

For the paintings selected from $1948$, we observe an special case for \emph{(8)Summertime: Number 9A} with respect to the others.
This painting obtain the lowest fractal dimension of this group.
We analyzed in detail this behavior and we found that the local fractal dimension depends on the size or section taken in the image.

\begin{figure}
\begin{center}
 \includegraphics[width=0.95\textwidth]{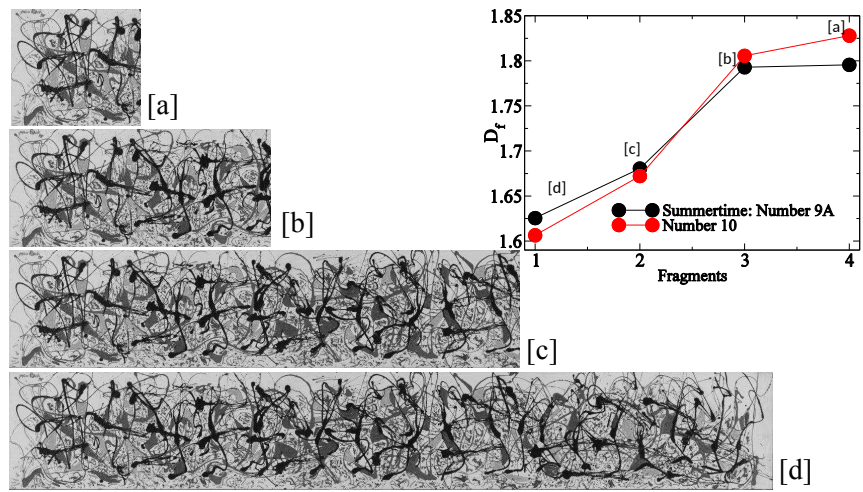}%
\caption{The figure shows the evolution of fractal dimension ($D_{f}$) for four different fragments of \emph{Summertime: Number 9A} and four different fragments of \emph{Number 10}. There are presented the fragments of \emph{Summertime: Number 9A}.
As the section of the image decreases from (d) to (a), the local fractal dimension increases. The same behavior occurs on this Pollock's paintings.\label{SP}}
\end{center}
\end{figure}

In Fig.(\ref{SP}) is shown the fractal evolution of \emph{(8)Summertime: Number 9A} and \emph{(14)Number 10} for different fragments taken from the whole original image.
We found that the paintings \emph{(8)Summertime: Number 9A} and \emph{(14)Number 10} reached a fractal dimension close to $1.6$ when is taken in to consideration the whole painting.
This indicates according to our approach that they are the paintings with paint strokes structured with major disorder than all.
We also found that, if we select a third part of the whole image and calculate the local fractal dimension using the same method, we obtained interest results reported in Fig.(\ref{SP}).
To do this, we divided the image into four fragments and we calculate its corresponding singularity spectrum for each part.

Fig.(\ref{SP}) shows that the Hausdorff-Besikovich fractal dimension increases in an inversely way as the size of the image.
This could be interpreted as an evidence of self-similarity in Pollock's paintings and the effect of the scale of observation.
The same treatment was made on \emph{(14)Number 10} and we obtain the same behavior that {(8)Summertime: Number 9A}.
It is interesting that we did not find this behavior for all the others selected Pollock's paintings, just on this two cases.

According to our analysis, the group of paintings from 1948: \emph{(13)Number 14, Gray}, on Fig.(\ref{FPA}) present the highest fractal dimension.
This can be correlated to an homogeneous distribution of the paint strokes.

Meanwhile, \emph{(15)Autumn Rhythm} reaches the minimum $D_{f}$ of the $1950$ paintings group, i.e it was the painting with a disorder distribution of paint strokes.
On the other hand \emph{(18)Number 18} presents the highest degree of order with a fractal dimension close to two.

For the $1951$ paintings group, \emph{(22)Untitled} presents an homogeneous paint strokes distribution.
This can be correlated to its higher local fractal dimension.
The opposite case is represented by \emph{(19)In Echo: Number 25}.
This group of paintings present a dramatic evolution of fractality.
This can be interpreted as the dripping technique was controlled and directed by Pollock in a more perfect way.
These last four canvases were painted in the last years of Pollock's life, when he improved and know perfectly his famous technique.

The hypothesis that fractal dimension increases as a function of the year in which the paintings were painted, was tested in this report.
We found effectively this kind of behavior.
\emph{Untitled} painted in $1951$ was the one with paint strokes distributed in a more ordered structure, according to our interpretation, while \emph{(2)Lucifer} was the most fractal one, without considering the peculiar cases of \emph{(8)Summertime: Number 9A} and \emph{(14)Number 10}.

\section{Authentication}
The fractal dimension has been questioned as an order parameter to authenticate abstract artworks~\cite{MureikaChaos,Jones,Jones1}.
However, it is unquestionable its use to characterize digital images from physical systems~\cite{Barnsley,Calleja1,Cervantes,Carrillo2}.
Why it can not be used to characterized abstract art?
We believe that Jackson Pollock developed a unique painting technique, and his creations are fantastic and unrepeatable.
Evidently they present certain degree of complexity, however, it has been measured.
Obviously the authentication processes require more than one parameter, however the fractal dimension seems to be adequate for his kind of artworks~\cite{Shamir,Irfan,Hany}.

\begin{figure}
\begin{center}
 \includegraphics[width=0.98\textwidth]{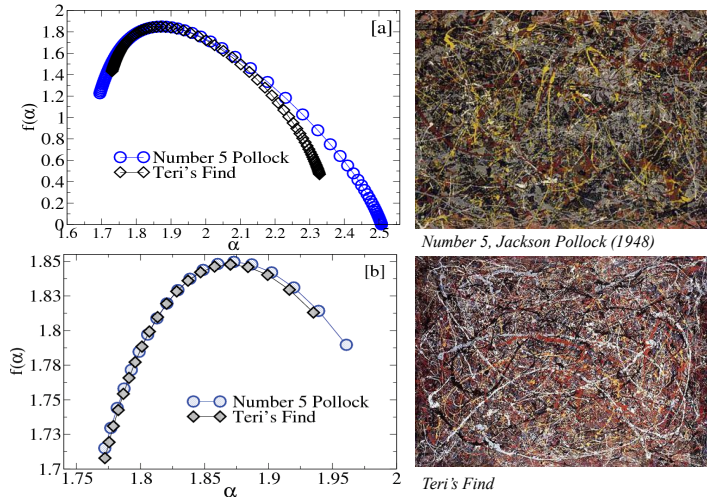}
\caption{The figure shows the multi-fractal spectrum of the called \emph{Teri's Find} and the Pollock \emph{Number 5}. The spectrums are too similar between them.\label{teri}}
\end{center}
\end{figure}

We obtained the multi-fractal spectrum of the painting called \emph{Teri's Find}~\cite{Teri,Teri2}.
Fig.(\ref{teri})\emph{(a)} is shown the complete multi-fractal spectrum of \emph{Teri's Find} not attributed to Pollock and \emph{Number 5} recognized as painted by Jackson Pollock.
We obtain that both spectrums look very similar.
The unique difference is that the Pollock's painting presents a longer multi-fractal spectrum than the\emph{Ter's Find}.
However they agree on many of their values.

Fig.(\ref{teri})\emph{(b)} shows details on the maximum values of both multi-fractal spectrums.
\emph{Teri's Find} reaches $D_{f}=1.8477$ and \emph{Number 5} obtains $D_{f}=1.8496$.
Quantitatively also both spectrums are very close.
Could be possible that \emph{Teri's Find} was painted by Pollock based on the curved of its multi-fractal dimension? or It is the fractal dimension insufficient to quantify abstract art?

According to our results \emph{Teri's Find} presents a high degree of complexity in a similar way than Pollock's paintings are, and taking into account the magnitude of its value, it is within the range of characteristic values found on Pollock's artworks. Due to that, we support the idea that \emph{Teri's Find} was painted by Jackson Pollock.

\section{Conclusions}
We report the multi-fractal behavior of twenty-two Pollock's paintings which were selected by considering the apparent complexity of the paint strokes, the year in which they were painted, the density of darkness strokes on top and the light background.

The left side of the spectrums is presented and was found that all the paintings have many local fractal dimensions.
This suggest that the magnitude of fractal dimension depends on the scale of observation ($\varepsilon$).
The length of all the left side of the spectrums grows from $\alpha=1.5$ to $\alpha=2$ with the notable exceptions of \emph{(8) Summertime: Number 9A} and \emph{(14) Number 10} which begins from $\alpha=1.5$.
This result corroborates the self-similarity of the Pollock's paintings.

The visual complexity of the paintings was quantified by the maximum value of the multi-fractal spectrum as a function of the year in which the paintings were painted.
We obtain a range of values between $D_{f}=1.78$ up to $D_{f}=1.88$.
From the definition of order using in this report, the degree of order in Pollock's paintings grows as a function of the year in which they were painted, taking into account the sequence of paintings listed in Table (\ref{table-1}).
This result could be an indication as a perfect knowledge of the dripping technique by the artist, a manifestation of control and perfection of his dripping technique and we can suggest that paint strokes were made consciously.

The self-similarity was tested on \emph{(8)Summertime:Number 9A} and \emph{(14)Number 10}.
The results reported in Fig.(\ref{SP}) indicate that fractal dimension increases inversely as a function of the size of the digital image.

Finally, \emph{Teri's Find} was tested by our method, and we found a similar multi-fractal spectrum between this not recognized painting, that could have been painted by Pollock and \emph{Number 5}, painted by Pollock.
We suggest that fractal dimension can not be definitely rejected as a parameter to authenticate abstract artworks.
Despite, this statement has been tested by different methods.

Many aspects have to be consider to analyze Pollock's paintings, such as the right kind of paint and brush, the correct speed and movements not only of the hand even also the arm; all in exactly concordance with the knowledges and the individual projection of the artist.
For all of that, the description of Pollock's paintings needs to be made taking into account a combination of different scientific techniques, and fractal dimension can be a good quantitative parameter without invasive techniques.



\end{document}